\documentclass[letterpaper, 10 pt, conference]{ieeeconf}  

\IEEEoverridecommandlockouts                              

\usepackage{graphics} 
\usepackage{epsfig} 
\usepackage{mathptmx} 
\usepackage{times} 
\usepackage{amsmath} 
\usepackage{amssymb}  

\usepackage{mathtools}
\usepackage{subfigure,rotating}
\usepackage{amsmath,amssymb,amsfonts}
\usepackage{algorithmic}
\usepackage{graphicx,subfigure}
\usepackage{textcomp}
\usepackage{wrapfig}
\usepackage{gensymb}
\usepackage{bm}
\usepackage{float}
\usepackage{color,soul}
\usepackage{array}
\usepackage{multirow}
\usepackage{array}
\usepackage[acronym]{glossaries}
\usepackage{mdframed}
\usepackage{fancyvrb}
\usepackage[utf8]{inputenc}
\usepackage{algorithm}
\usepackage{hyperref}
\usepackage[hyphenbreaks]{breakurl}

\newcommand{\V}[1] {\mathbf{#1}}

\newcommand{\dabs}[1] {\left \| #1 \right \|}
\newcommand{\lln}[1] {\dabs{#1}_2}

\newcommand{\beq} {\begin{equation}}
\newcommand{\eeq} {\end{equation}}
\newcommand{\beqn} {\begin{eqnarray}}
\newcommand{\eeqn} {\end{eqnarray}}

\newcommand{\bc}{\begin{center}}
\newcommand{\ec}{\end{center}}
\newcommand{\bi}{\begin{itemize}}
\newcommand{\ei}{\end{itemize}}
\newcommand{\be}{\begin{enumerate}}
\newcommand{\ee}{\end{enumerate}}

\newacronym{fov}{FOV}{field of view}
\newacronym{aov}{AOV}{angle of view}
\newacronym{loc}{LOC}{level of confidence}
\newacronym{pod}{POD}{probability of detection}
\newacronym{rbe}{RBE}{recursive Bayesian estimation}
\newacronym{pdf}{PDF}{probability density function}
\newacronym{pt}{PT}{pan/tilt}
\newacronym{uav}{UAV}{unmanned aerial vehicle}
\newacronym{ugv}{UGV}{unmanned ground  vehicle}
\newacronym{1d}{1D}{one-dimensional}
\newacronym{2d}{2D}{two-dimensional}
\newacronym{3d}{3D}{three-dimensional}
\newacronym{kf}{KF}{Kalman filter}
\newacronym{ekf}{EKF}{extended Kalman filter}
\newacronym{pf}{PF}{particle filter}
\newacronym{slam}{SLAM}{Simultaneous Localization and Mapping}
\newacronym{mbzirc}{MBZIRC}{Mohamed Bin Zayed International Robotics Challenge}
\newacronym{gps}{GPS}{Global Positioning System}
\newacronym{imu}{IMU}{Inertial Measurement Unit}
\newacronym{ros}{ROS}{Robot Operating System}
\newacronym{com}{COM}{center of mass}
\newacronym{dt}{DT}{distance transform}
\newacronym{ooi}{OOI}{object of interest}
\newacronym{sfm}{SfM}{Structure from Motion}
\newacronym{doe}{DOE}{Design of Experiments}
\newacronym{v&v}{V\&V}{Verification and Validation}
\newacronym{de}{DE}{Differential Entropy}
\newacronym{sar}{SAR}{search and rescue}
\newacronym{ba}{BA}{bundle adjustment}
\newacronym{ndt}{NDT}{normal distribution transform}
\newacronym{gf2mm}{GF2MM}{grid feature to map matching}
\newacronym{ogm}{OGM}{occupancy grid map}
\newacronym{udm}{UDM}{unoccupancy distance map}
\newacronym{tsp}{TSP}{travelling salesman problem}
\newacronym{bim}{BIM}{building information model}
\newacronym{ml}{ML}{machine learning}
\newacronym{icp}{ICP}{Iterative Closest Point}
\newacronym{dnn}{DNN}{deep neural network}
\newacronym{rrt}{RRT}{rapidly-exploring random trees}
\newacronym{ik}{IK}{inverse kinematics}
\newacronym{fk}{FK}{forward kinematics}
\newacronym{rmse}{RMSE}{root mean square error}

\DeclareMathOperator*{\argmin}{argmin}
\DeclareMathOperator*{\Log}{Log}

\usepackage[dvipsnames]{xcolor}

\title{\LARGE \bf
Lie Theory Based Optimization for Unified State Planning of Mobile Manipulators
}

\author{William Smith$^{1}$, Siddharth Singh$^{2}$, Julia Rudy$^{3}$, and Yuxiang Guan$^{4}$
\thanks{$^{1}$Analytical and Spatial Intelligence Research Team,
        Applied Research Associates, Raleigh NC, 27612, USA
        {\tt\small wbs3ra@virginia.edu}}%
\thanks{$^{2}$Department of Mechanical and Aerospace Engineering,
        University of Virginia, Charlottesville VA, 22903, USA
        {\tt\small sks4zk@virginia.edu}}%
\thanks{$^{3}$Department of Electrical and Computer Engineering,
        University of Virginia, Charlottesville VA, 22903, USA
        {\tt\small jr7yj@virginia.edu}}%
\thanks{$^{4}$Department of Mechanical Engineering,
        The University of Texas at Dallas, Richardson TX, 75080, USA
        {\tt\small yuxiang.guan@utdallas.edu}}%
}

\begin{document}

\maketitle
\thispagestyle{empty}
\pagestyle{empty}

\begin{abstract}
Mobile manipulators are finding use in numerous practical applications. 
The current issues with mobile manipulation are the large state space owing to the mobile base and the challenge of modeling high degree of freedom systems. 
It is critical to devise fast and accurate algorithms that generate smooth motion plans for such mobile manipulators. 
Existing techniques attempt to solve this problem but focus on separating the motion of the base and manipulator. 
We propose an approach using Lie theory to find the inverse kinematic constraints by converting the kinematic model, created using screw coordinates, between its Lie group and vector representation. 
An optimization function is devised to solve for the desired joint states of the entire mobile manipulator. 
This allows the motion of the mobile base and manipulator to be planned and applied in unison resulting in a smooth and accurate motion plan. 
The performance of the proposed state planner is validated on simulated mobile manipulators in an analytical experiment. 
Our solver is available with further derivations and results at \url{https://github.com/peleito/slithers}.
\end{abstract}

\begin{keywords}
mobile manipulation, Lie theory, state planning
\end{keywords}

\section{Introduction}
\label{sec:intro}

Mobile manipulation is a fundamental task in robotics with valuable applications in various fields. 
Mobile manipulators were first used for exploration in remote environments on the surface of celestial bodies~\cite{putz_space_1998}. 
In recent decades, with the proliferation of robotic systems, mobile manipulators have found use in several domains to aid human workers such as construction~\cite{MM_const}, agriculture~\cite{MM_agri}, additive manufacturing ~\cite{MM_addmfg}, telehealth~\cite{MM_nurse}, mapping~\cite{smith_mapping_2022,smith_mapping_2023}, etc. 
Mobile manipulators have also been used in areas that humans can not safely access in disaster related tasks by observing the scene to create three dimensional environments~\cite{schwarz_nimbo_2017} or measuring environmental conditions, such as radiation~\cite{nagatani_emergency_2013}. 
New companies have even been formed to push mobile manipulators to low level consumers in offices or homes~\cite{kalashnikov_deep_2021}. 
Due to the many use cases and immense potential of mobile manipulators, it is imperative to provide solutions for state planning to aid in its practical use.  

A common problem for all mobile manipulation tasks is maneuvering the mobile base and the manipulator in unison to complete a task. 
Attempts have been made to integrate a robotic manipulator with a mobile platform, but the motion of two robots is typically controlled separately~\cite{braun_rhh-lgp_2022} leading to sporadic movement of the entire robotic system. 
This paper proposes the motion of a robotic manipulator and mobile platform be controlled in unison, increasing the usability of mobile manipulators in practical applications. 

\subsection{Previous Work}

Existing methods of state planning for mobile manipulators can be broadly classified into two approaches. 
The first approaches comprise kinematic based planning methods that solve all the joint states based on a given pose goal. 
The goals are converted into joint states using model based inverse kinematics~\cite{muzan_implementation_2012} or optimization based inverse kinematics~\cite{schappler_euler_2019}. 
Due to the under constrained nature of the problem, many different techniques have been pursued leveraging quadratic optimization~\cite{maric_inverse_2020}, evolutionary algorithms~\cite{berenson_optimization_2008}, and constrained inverse kinematics~\cite{pardi_path_2020}. 
These methods utilize conventional planning methods~\cite{lavalle_randomized_1999} for interpolating between the states in the trajectory to create a finely sampled path in state space. 
Such methods are appropriate for planning mobile manipulator states with low degrees of freedom but have limited scalability. 
Therefore these methods are not suited for tasks requiring complex manipulation, such as inspection of large infrastructure~\cite{macaulay_machine_2022}.

In the second approaches, the motion of mobile manipulators and bases are solved separately~\cite{yu_base_2018}, leveraging traditional planning methods. 
The mobile base is maneuvered towards regions to maximize the manipulability for the respective task~\cite{Colucci_kinematic_2022}. 
This leads to a discretized path with frequent stops of the base while the manipulator actuates to the limits of the reachable space~\cite{megalingam_autonomous_2020}. 
Many methods aim to determine the optimal positions for the mobile platform by minimizing joint motion~\cite{sarker_screw_2020} and minimizing base motion~\cite{paus_combined_2017}. 
These methods then use conventional inverse kinematics~\cite{trutman_inverse_2020} to actuate the manipulator. 
Since these techniques mostly focus on moving from one desired state to another and do not consider the continuous motion along the path, it is difficult for the robot to follow long paths smoothly. 
Consequently, these methods are unsuitable for problems requiring mobile manipulation in large work environments, such as urban search and rescue~\cite{wang_development_2023}. 

\subsection{Our Contributions}

This paper presents SLITHERS (State planner using LIe THEory for RoboticS), a Lie theory based optimization approach for generating smooth, accurate, and unified motion plans. 
Given a sequence of desired end effector poses, the proposed method solves for the joint values using kinematic constraints computed leveraging Lie theory. 
The joint state values are estimated by solving a constrained optimization problem which yields the sequence of states given the current pose and desired pose. 
In this study, our contributions are three-fold: 
1) firstly, we successfully developed a new motion planner given unrestricted motion for a high degrees of freedom robot (8-10 degrees of freedom), 
2) the proposed method is generalizable and easily adjustable for mobile manipulators of differing degrees of freedom, 
3) and lastly the proposed method guarantees unified motion between the base and manipulator. 
The proposed method is validated based on planning accuracy and other metrics in a simulated experiment. 
During validation, it was found that our method achieves significantly high accuracy and considerable smoothness.

This paper is organized as follows. 
Section~\ref{sec:background} describes the forward kinematics of a manipulator using screw coordinates and Lie theory.  
Section~\ref{sec:proposed} presents SLITHERS, the optimization based state planner for mobile manipulators. 
Experimental validation is presented in Section~\ref{sec:results}, and Section~\ref{sec:conclusions} summarizes conclusions and ongoing work.  

\section{Manipulator Kinematics}
\label{sec:background}

It is essential to formulate suitable kinematic models for robotic systems to analyze the behavior of robot manipulators. 
To develop a kinematic model of the mobile manipulator, the physical parameters of the robot must be measured.
The Denavit-Hartenberg (DH)~\cite{hartenberg_kinematic_1964} method utilizes four parameters and is the most common method for deriving the homogeneous transformation matrices of a robotic manipulator. 
Even though the process for determining the kinematics of a manipulator is simplified through the DH method, the process requires that a strict convention be followed when assigning body fixed frames to all of the joints. 
By contrast, Lie theory's generic format makes it easier to model mobile manipulator kinematics and dynamics for joints with unrestricted motion, yielding better planning and control. 
Since a mobile manipulator consists of both a manipulator and a mobile base, the forward kinematics must be calculated for both the base and manipulator. 
In this section, Lie theory based forward kinematics are shown for constructing the kinematic model of a manipulator. 

\subsection{Screw Coordinates}
\label{sec:screw}

Conventionally, homogeneous transformations found using DH parameters are used to represent the relationship between two robot links. 
Instead, screw coordinates can be used to define the motion for all of the joints in a robotic manipulator. 
Screw theory~\cite{Ball1876} helps define the position and motion of rigid bodies with respect to a fixed reference frame by classifying all motion as a screw. 
This allows for a generic method to be applied to all motions regardless of the degrees of freedom associated with the motion. 
Body fixed frames are not necessary to be defined as only one frame representing the fixed reference frame is necessary. 
A twist, $\mathbf{S} \in \mathbb{R}^6$, is a screw composed of two three dimensional vectors. 
It represents the rotational, $\mathbf{S}_{\omega}$, and translational, $\mathbf{S}_{v}$, motion about an axis as given by
\begin{equation}
    \mathbf{S} =    \begin{Bmatrix}
                        \boldsymbol{\omega} \\
                        \mathbf{v} + \mathbf{d} \times \boldsymbol{\omega}
                        \end{Bmatrix} \, ,
\end{equation}
where $\boldsymbol{\omega} \in \mathbb{R}^3$ is the angular motion, $\mathbf{v} \in \mathbb{R}^3$ is the linear motion, and $\mathbf{d} \in \mathbb{R}^3$ is a translation from the reference frame to the joint. 
Since screw joints can be used to represent all of the joints of a mobile manipulator, screw theory can be used to represent the kinematic relationships of the entire robot.

Screw theory is useful for forward and inverse kinematics since revolute and prismatic joints are represented in the same format.
Since most actuators only provide one degree of freedom, many systems can be divided into a set of revolute and prismatic joints corresponding to the total motion.
In a revolute joint ($\mathbf{v}=\mathbf{0}$), the twist element, $\mathbf{S}_{\omega}$, is defined by rotational motion and translational vectors as given by 
\begin{equation}
    \mathbf{S}_{\omega} =    \begin{Bmatrix}
                        \boldsymbol{\omega} \\
                        \mathbf{d} \times \boldsymbol{\omega}
                        \end{Bmatrix} \, .
\end{equation}
A prismatic joint ($\boldsymbol{\omega}=\mathbf{0}$) has a twist element, $\mathbf{S}_{v}$, defined by the linear motion vector as given by
\begin{equation}
    \mathbf{S}_{v} =    \begin{Bmatrix}
                        \mathbf{0} \\
                        \mathbf{v}
                        \end{Bmatrix} \, .
\end{equation}
The defined screw coordinates can then be used to compute the forward and inverse kinematics from the fundamentals of Lie theory using the product of exponentials. 

\subsection{Lie Groups and Lie Algebra}
\label{sec:lie_forward}

\begin{figure}[tbh]
\centerline {\includegraphics[width=\linewidth/4*3]{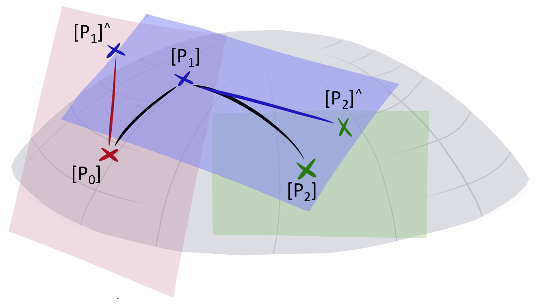}}
\caption{Pictorial representation of using Lie theory to move across the Lie group using the corresponding Lie algebra. The motion in between each point on the Lie group ($[P_{n}]$) can be represented as a motion on the corresponding Lie algebra ($[P_{n}]^{\wedge}$).} \label{fig:lie_theory}
\end{figure}

Lie theory is the foundation that relates the Lie group and its corresponding Lie algebra through the exponential map. 
Lie groups are special smooth manifolds commonly found in the field of robotics~\cite{micro_lie}. 
Typical groups in robotics are for rotations defined by the special orthogonal group, $SO(n)$, and rigid motion defined by the special Euclidean group, $SE(n)$.
Through the matrix exponential, the motion at a given time can transform the current pose into the next pose, as seen in Figure~\ref{fig:lie_theory}.  
The matrix exponential directly maps the motion from the Lie algebra to the Lie group enabling fundamental mathematical operations to be performed in the tangent space and then mapped back to the group. 
One advantage of performing planning on the Lie algebra is the ease of finding the shortest path between points as compared to solving for a manifold. 
On the Lie algebra, the shortest path is the line connecting the two points which represents the geodesic when mapped back to the Lie group.  
In the case of a three dimensional rigid body transformation, $[\mathbf{P}] \in SE(3)$, any element in the Lie group can be defined by a rotation matrix, $[\mathbf{R}]$, and translation vector, $\mathbf{t}$, as given by
\begin{equation}
    [\mathbf{P}] = \begin{bmatrix}
[\mathbf{R}] & \V{t} \\
\mathbf{0} & 1
\end{bmatrix} \, .
\end{equation}
The corresponding Lie algebra can be found by creating a tangent space on the group manifold at the group identity. 
Due to the properties of a Lie group, the tangent space at any point on the manifold is identical and therefore the Lie algebra can be used to represent the tangent space of any point on the manifold. 
The Lie algebra, $[\mathbf{S}]^{\wedge} \in \mathfrak{se}(3)$, for a homogeneous transformation is defined by the twist governing the motion and given by
\begin{equation}
    [\mathbf{S}]^{\wedge} = \begin{bmatrix}
[\boldsymbol{\omega}]_{\times} & \mathbf{v} + \mathbf{d} \times \boldsymbol{\omega} \\
\mathbf{0} & 0
\end{bmatrix} \, ,
\end{equation}
where $[\cdot]_{\times}$ indicates a skew symmetric matrix.

Lie theory is especially helpful for forward kinematics of robotic manipulators as joints are broken down into purely revolute and prismatic joints. 
The matrix exponential is used to map $[\mathbf{S}(q)]^{\wedge}$ onto the $SE(3)$ manifold changing the pose from $[\mathbf{P}_{0}]$ to $[\mathbf{P}]$ as given by
\begin{equation}
    [\mathbf{P}] = e^{[\mathbf{S(q)}]^{\wedge}} [\mathbf{P}_{0}] \, ,
\end{equation}
where $[\mathbf{P}_{0}]$ is the configuration state at the identity and $S$ is a function of $q$, the joint state.
Since the twist coordinates are defined with respect to a fixed reference frame, right hand multiplying $[\mathbf{P}_{0}]$ ensures $[\mathbf{S}(q)]^{\wedge}$ is time invariant. 
When there is a kinematic chain of rigid links as in a manipulator, the product of exponentials is used to solve for the forward kinematics along the entire chain as given by
\begin{equation}
\label{eq:product_of_exp}
    [\textbf{P}_{n}^{0}] = \prod_{i \in n} \left( e^{[\mathbf{S}_{i}(q_{i})]^{\wedge}} \right) [\mathbf{P}_{0}] \, , 
\end{equation}
where $n$ is the number of links in the robotic system. 
It can be seen that the forward kinematics takes a similar form as compared to using DH parameters but allows for all motion, including the motion of the mobile base, to be modeled in the same manner. 

\section{State Planning using Lie Theory Optimization}
\label{sec:proposed}

\begin{figure}[tbh]
\centerline {\includegraphics[width=\linewidth]{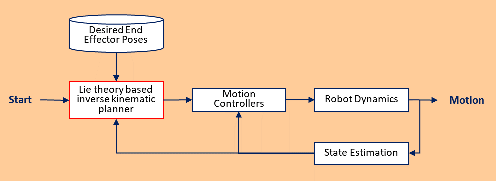}}
\caption{The framework of the proposed method takes in a set of desired end effector poses, computes the desired state based on an objective function and executes the motion to achieve the desired states.} \label{fig:proposed}
\end{figure}

Let us assume the desired path of the end effector is specified and given by $\mathbf{\mathcal{P}}^{*}_{0:T} = \{ [\mathbf{P}^{w}_{n}]^{*}(0),[\mathbf{P}^{w}_{n}]^{*}(\Delta t), \cdots , [\mathbf{P}^{w}_{n}]^{*}(T) \}$ where [$\mathbf{P}^{w}_{n}](t) \in SE(3)$ is the pose of the end effector at any given time. 
The time sampling duration is represented by $\Delta t$ and the total time duration is represented by $T$. 
At any given time the state of the robot is defined as $\mathbf{x}_{r} = [ \mathbf{q}, \mathbf{\dot{q}}, \mathbf{t}_{r}, \mathbf{v}_{r}, \boldsymbol{\theta}_{r}, \boldsymbol{\omega}_{r} ]^\top$. 
Here $\mathbf{q} = [\theta_1, \theta_2, \cdots, \theta_n] \in \mathbb{R}^n$ represents the state vector of the manipulator in joint space and consequently, $\mathbf{\dot{q}}$ represents the joint velocity vector. 
The mobile base position in the world coordinates is represented by $\mathbf{t} \in \mathbb{R}^3$ and $\mathbf{v} \in \mathbb{R}^3$ represents the mobile base velocity. 
The global orientation and global angular velocity of the mobile base are represented by $\boldsymbol{\theta}_r$ and $\boldsymbol{\omega}_r$, respectively. 
Figure~\ref{fig:proposed} shows the basic schematic for the proposed state planning method. 
The current state of the robot is assumed to be known with minimal uncertainty from joint encoders and state estimation of the base. 
Onboard motion controllers will then take the desired state to determine the low level input based on the dynamical model. 
After the elapsed time, the desired states will be computed again until all of the desired poses are reached. 

A proper inverse kinematics solver should be grounded in a strong forward kinematic model of the robot. 
Full derivations for the forward and inverse kinematics of the robotic models used can be found at \url{https://github.com/peleito/slithers}.
Since the inverse kinematics of a mobile manipulator are rank deficient, the forward kinematics can be used as the basis for an optimization problem. 
Analytical and numerical methods are the two main techniques used to solve inverse kinematics problems. 
The joint variables can be calculated analytically based on given configuration data but there is not always a solution due to the rank of the resulting system of equations. 
Therefore, a numerical method is pursued to estimate the joint variables required to achieve the desired state. 
Several optimization methods could be utilized in state planning, such as $minimum$ $time$, $minimum$ $energy$, and $minimum$ $jerk$. 
In the following subsection, an optimization method to achieve $minimum$ $jerk$ is proposed to achieve a desired pose while maintaining the smooth motion of the base and joints. 

\subsection{Inverse Kinematic Constraints Using Lie Theory}
\label{sec:lie_inverse}

Lie theory and screw coordinates are useful for computing the inverse kinematic constraints since it uses the change in states which is naturally suited for executing planned states. 
It also takes advantage of connecting poses along the geodesic of the manifold, reducing major unnecessary motion. 
The transition from forward kinematics to inverse kinematics is intuitive since the state variables can be easily separated from the pose variables. 

First, the forward kinematics of the robotic system are represented using the product of exponentials as seen in Equation~\ref{eq:product_of_exp}, but are adapted to consider a time series of motion by introducing time steps denoted by the subscript $k$. 

The screw coordinates of the manipulator, $\mathbf{S}_i \in \mathbb{R}^6$, and configuration state at the identity, $[\mathbf{P}_{0}] \in SE(3)$, are defined with respect to a body fixed frame on the base of the robot to ensure they are time invariant. 
The entire pose is then projected into the world frame through the transformation from the base frame to the world frame, $[\mathbf{P}_{r,k}^{w}]$. 
The forward kinematics to the next pose, $[\mathbf{P}_{n,k+1}^{w}]^{*}$, are now given by
\begin{multline}
    [\mathbf{P}_{n,k+1}^{w}]^{*} = [\mathbf{P}_{r,k}^{w}] e^{[\mathbf{S}_{r}(\mathbf{v}_{r,k+1}, \boldsymbol{\omega}_{r,k+1})]^{\wedge}} \\
    * \ \prod_{i \in n} \left( e^{[\mathbf{S}_{i}(q_{i,k+1})]^{\wedge}} \right) [\mathbf{P}_{0}] \, , 
\end{multline}
where $\mathbf{S}_{r}$ is the screw coordinates for the robotic base. 
In the case of a mobile manipulator, the Lie algebra, $[\mathbf{S}_{r}(\mathbf{v}_{r,k+1}, \boldsymbol{\omega}_{r,k+1})]^{\wedge}$, represents the degrees of freedom provided by the mobile platform. 
Common mobile platform configurations provide one to three degrees of freedom in addition to the degrees of freedom granted by the manipulator.

The inverse kinematics can be solved by isolating the joint states from the poses to yield
\begin{multline}
    [\mathbf{P}_{r,k}^{w}]^{-1}[\mathbf{P}_{n,k+1}^{w}]^{*}[\mathbf{P}_{0}]^{-1} = \\
    e^{[\mathbf{S}_{r}(\mathbf{v}_{r,k+1}, \boldsymbol{\omega}_{r,k+1})]^{\wedge}} \prod_{i \in n} \left( e^{[\mathbf{S}_{i}(q_{i,k+1})]^{\wedge}} \right) \, .
\end{multline}
The resulting system is complex to solve and difficult to isolate the state variables using basic linear algebra. 
Therefore, the matrices are mapped to vector space using the $\Log$ operator~\cite{micro_lie} as given by
\begin{equation}
    \boldsymbol{\tau} = \Log\left([\mathbf{P}]\right) = \log\left([\mathbf{P}]\right)^{\vee}  \, , 
\end{equation}
where $\boldsymbol{\tau} \in \mathbb{R}^6$ is the screw coordinate representation comprising the rotation and translation between poses. 
The $\Log$ operator maps between the Lie group manifold and vector space. 
The entities on the manifold are first mapped to the tangent space with the matrix logarithm, $\log(\cdot)$. 
They are then mapped from the tangent space to the vector space with the vee operator, $(\cdot)^{\vee}$. 

Therefore the inverse kinematic constraints in vector form are denoted by
\begin{multline}
\label{eq:inverse_equation}
    \Log\left([\mathbf{P}_{r,k}^{w}]^{-1}[\mathbf{P}_{n,k+1}^{w}]^{*}[\mathbf{P}_{0}]^{-1}\right) = \\
    \Log\left(e^{[\mathbf{S}_{r}(\mathbf{v}_{r,k+1}, \boldsymbol{\omega}_{r,k+1})]^{\wedge}} \prod_{i \in n} \left( e^{[\mathbf{S}_{i}(q_{i,k+1})]^{\wedge}}\right) \right) \, .
\end{multline}
For simplification purposes let 
\begin{equation}
    \boldsymbol{\tau}_{poses}\left([\mathbf{P}_{n,k+1}^{w}]^{*}\right) = \Log\left([\mathbf{P}_{r,k}^{w}]^{-1}[\mathbf{P}_{n,k+1}^{w}]^{*}[\mathbf{P}_{0}]^{-1}\right) \, ,
\end{equation}
where $\boldsymbol{\tau}_{poses}\left([\mathbf{P}_{n,k+1}^{w}]^{*}\right)$ represents the pose transformations and let 
\begin{multline}
    \boldsymbol{\tau}_{joints}\left(\mathbf{u}_{k+1}\right) = \\
    \Log\left(e^{[\mathbf{S}_{r}(\mathbf{v}_{r,k+1}, \boldsymbol{\omega}_{r,k+1})]^{\wedge}} \prod_{i \in n} \left( e^{[\mathbf{S}_{i}(q_{i,k+1})]^{\wedge}} \right) \right) \, , 
\end{multline}
where $\boldsymbol{\tau}_{joints}\left(\mathbf{u}_{k+1}\right)$ represents the joint transformations in vector form and $\mathbf{u}_{k+1} = \{\V{q}_{k+1}, \mathbf{v}_{r,k+1}, \boldsymbol{\omega}_{r,k+1}\}$ is the input vector for the robotic system. 

\subsection{Optimization Based Inverse Kinematics}
\label{sec:optimization}

The optimization problem is formulated to solve for the joint variables of the mobile manipulator, $\mathbf{u}^{*}_{k+1}$, derived from the inverse kinematics in Equation~\ref{eq:inverse_equation}. 
The pose goals can be achieved by setting the pose error as the cost function.
Other constraints can be added to create and adapt the objective function to the robot's specific task. 
Since different tasks expect different performance from the mobile manipulator, it is best to start with the generic formulation. 
The basic formulation of the optimization problem is given by
\begin{equation}
\label{eq:basic_eq}
    \begin{aligned}
        \argmin_{\mathbf{u}^{*}_{k+1}} \quad & \mathrm{desired\ constraints}\\
        \textrm{s.t.} \quad     & \mathrm{required\ constraints} \, ,
    \end{aligned}
\end{equation}
where $\mathbf{u}^{*}_{k+1}$ is the optimization variable representing the next desired state of the mobile manipulator, `desired constraints' are constraints that can be afforded with an associated cost $\lambda$, and `required constraints' are constraints that can not be afforded. 
The state consists of the controllable parameters for the mobile robot including the linear and angular velocities of the mobile base, and the joint values of the manipulator.

As the optimization problem is made more specific for the task, the set of available states is reduced. 
The simplest formulation is minimizing the pose error between the next point in the path and the end effector by reformulating Equation~\ref{eq:inverse_equation} as a difference in the objective function. 
The motion is restricted by including joint limits to achieve the desired pose without violating joint constraints. 
The poses can be achieved by setting the pose error as a `desired constraint', which minimizes the predicted pose error between the forward kinematics of the end effector and the desired pose. 
Joint constraints are `required constraints' because the mobile manipulator can not physically occupy positions outside of the minimum and maximum joint values. 

For a simple task with no contact forces, the states can be planned to follow the poses in $\mathbf{\mathcal{P}}^{*}_{0:T}$ by 
\begin{equation}
    \begin{aligned}
        \argmin_{\mathbf{u}^{*}_{k+1}} \quad & \lln{\boldsymbol{\tau}_{joints}(\mathbf{u}^{*}_{k+1}) - \boldsymbol{\tau}_{poses}\left([\mathbf{P}_{n,k+1}^{w}]\right)} \\
        \textrm{s.t.} \quad     & \mathbf{u}^{*}_{k+1} \geq \V{u}_{\mathrm{min}}    \\
                                & \mathbf{u}^{*}_{k+1} \leq \V{u}_{\mathrm{max}} \, .
    \end{aligned}
\end{equation}
The `required constraints' are simply the joint limits given by $\V{u}_{\mathrm{min}}$ and $\V{u}_{\mathrm{max}}$. 
The current formulation can guarantee state smoothness between individual points, but to achieve continuous operation other constraints must be considered. 
An issue with such a basic formulation is the state smoothness of the joint solutions across the entire path, which is not currently considered. 
The pose error is likely to approach zero from the optimized joint values but the motion will not necessarily be feasible with many self collisions and joint discontinuities present. 

Since the time between points is given and the smoothness is of high priority, an optimization method to achieve \textit{minimum jerk} is proposed to plan states for following a path.
If \textit{minimum jerk} is achieved along each joint, the robot will maintain smooth motion of the mobile platform and manipulator.
Smoothness is ensured by adding two desired behaviors to the cost function, minimizing the motion and jerk of each joint. 
Minimizing the motion of each joint, $\mathbf{u}^{*}_{k+1}-\mathbf{u}_{k}$, increases the continuity of computed states and ensures the robot does not make any unnecessary movements between poses. 
The jerk of the joints, $\dddot{\mathbf{u}}^{*}_{k+1}$, is approximated by backwards finite differences and ensures the joints transition smoothly from one state to another along the length of the path. 

For the same task, the formulation is adapted to achieve smooth states by
\begin{equation}
    \begin{aligned}
        \argmin_{\mathbf{u}^{*}_{k+1}} \quad & \lambda_{e} \lln{\boldsymbol{\tau}_{joints}(\mathbf{u}^{*}_{k+1}) - \boldsymbol{\tau}_{poses}\left([\mathbf{P}_{n,k+1}^{w}]\right)}
         \\
         &+ \lln{\boldsymbol\lambda_v \odot \left(\mathbf{u}^{*}_{k+1}-\mathbf{u}_{k}\right)} + \lambda_j \lln{\dddot{\mathbf{u}}^{*}_{k+1}} \\
        \textrm{s.t.} \quad & \mathbf{u}^{*}_{k+1} \geq \V{u}_{\mathrm{min}}    \\
        & \mathbf{u}^{*}_{k+1} \leq \V{u}_{\mathrm{max}} \, ,
    \end{aligned}
\end{equation}
where $[\mathbf{A}]\odot[\mathbf{B}]$ is the Hadamard product and $\mathbf{u}_{k}$ is the current joint state. 
The new `desired constraints' are used for state smoothing, so the path is feasible for low level controllers to properly execute. 
The `required constraints' simply remain the joint limits, given by $\V{u}_{\mathrm{min}}$ and $\V{u}_{\mathrm{max}}$. 
The joint state continuity is controlled by $\lln{\boldsymbol\lambda_v \odot \left(\mathbf{u}^{*}_{k+1}-\mathbf{u}_{k}\right)}$ through limiting the motion of each joint in a weighted fashion. 
It is to be noted that $\lambda_e$, $\boldsymbol\lambda_v$, and $\lambda_j$ are scaling factors representing the relative importance. 

The goal pose is updated and the desired states are executed by low-level controllers to achieve the necessary motion to reach the desired states. 
The next states are to be solved in a step wise manner for the entire length of the path. 

\section{Experimental Validation}
\label{sec:results} 

The proposed method was tested on simulated mobile manipulators and assumed the system was fully controllable and observable with minimal sensor noise.
The planning performance of the proposed method was validated by accurately and smoothly following various end effector paths. 
The experiment was conducted to determine the feasibility of the proposed method as a generic and universal state planner by ensuring success on varying path types and various robotic systems. 

\begin{figure}[tbh]
\centerline {\includegraphics[width=\linewidth]{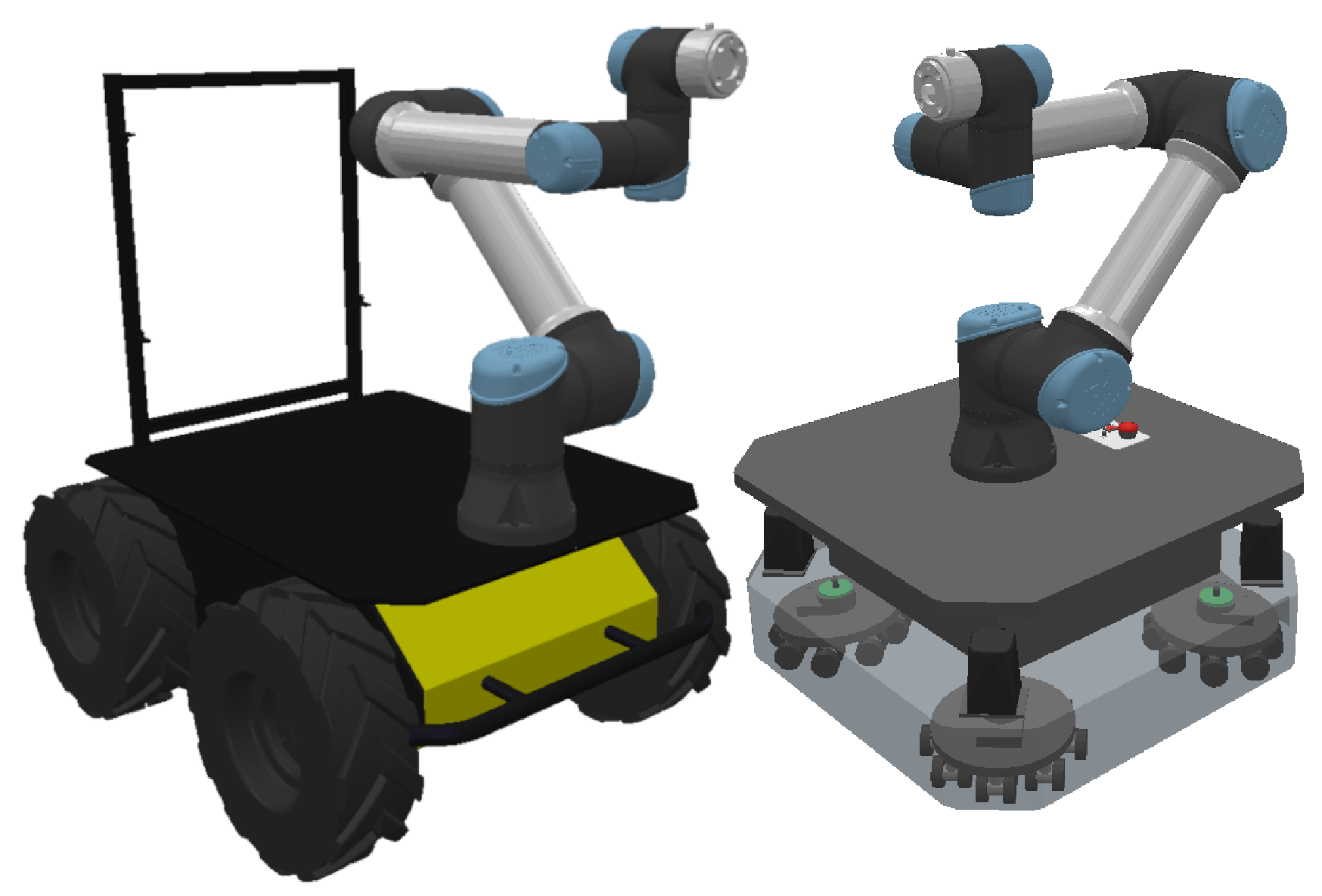}}
\caption{Mobile manipulators consisting of a six degree of freedom industrial robotic manipulator mounted on a non-holonomic mobile platform (left) and holonomic mobile platform (right).} \label{fig:robot_model}
\end{figure}

\begin{figure}[b]
\centerline {\includegraphics[width=\linewidth]{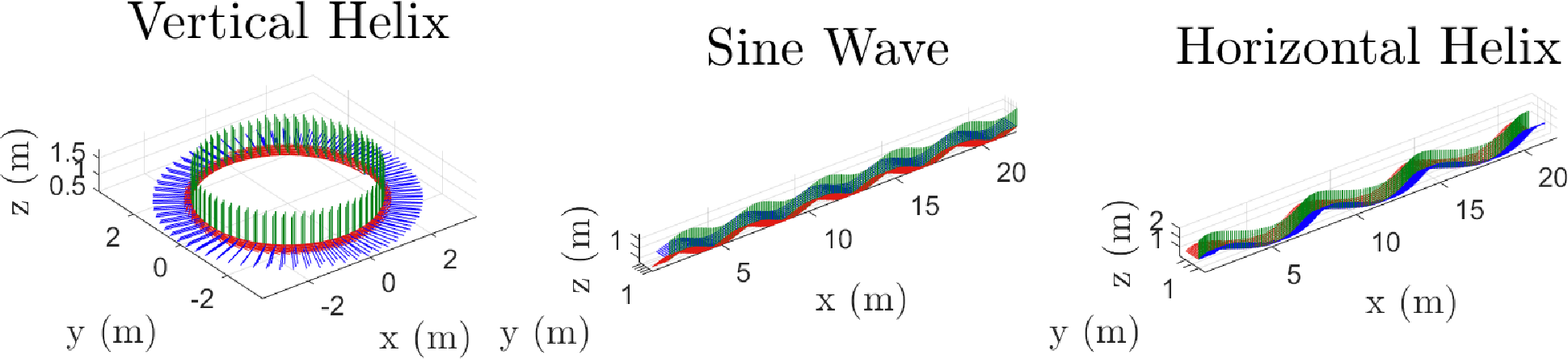}}
\caption{Test paths used for the simulated experiment with the colored axes representing the desired pose of the end effector. The red, blue, and green axes represent the x, y, and z axes, respectively.} \label{fig:full_poses}
\end{figure}

The robots tested were a six degree of freedom industrial collaborative manipulator (Universal Robots UR5e) on a non-holonomic mobile platform (Clearpath Robotics Husky) and a holonomic mobile platform (x-drive), as seen in Figure~\ref{fig:robot_model}. 
The test paths, shown in Figure~\ref{fig:full_poses}, have non-zero derivatives to test the continuity of states planned for differing path configurations. 
The parameters used for the experiment are $\Delta t = 0.2$ seconds, $T = 20$ seconds, $\lambda_e = 25$, $\lambda_j = 0.001$, $\boldsymbol\lambda_v = [1.0, 1.0, 0.25, 0.25, 0.1, 0.1, 0.1, 0.1]$. 

\begin{figure}[tbh]
\centerline {\includegraphics[width=\linewidth]{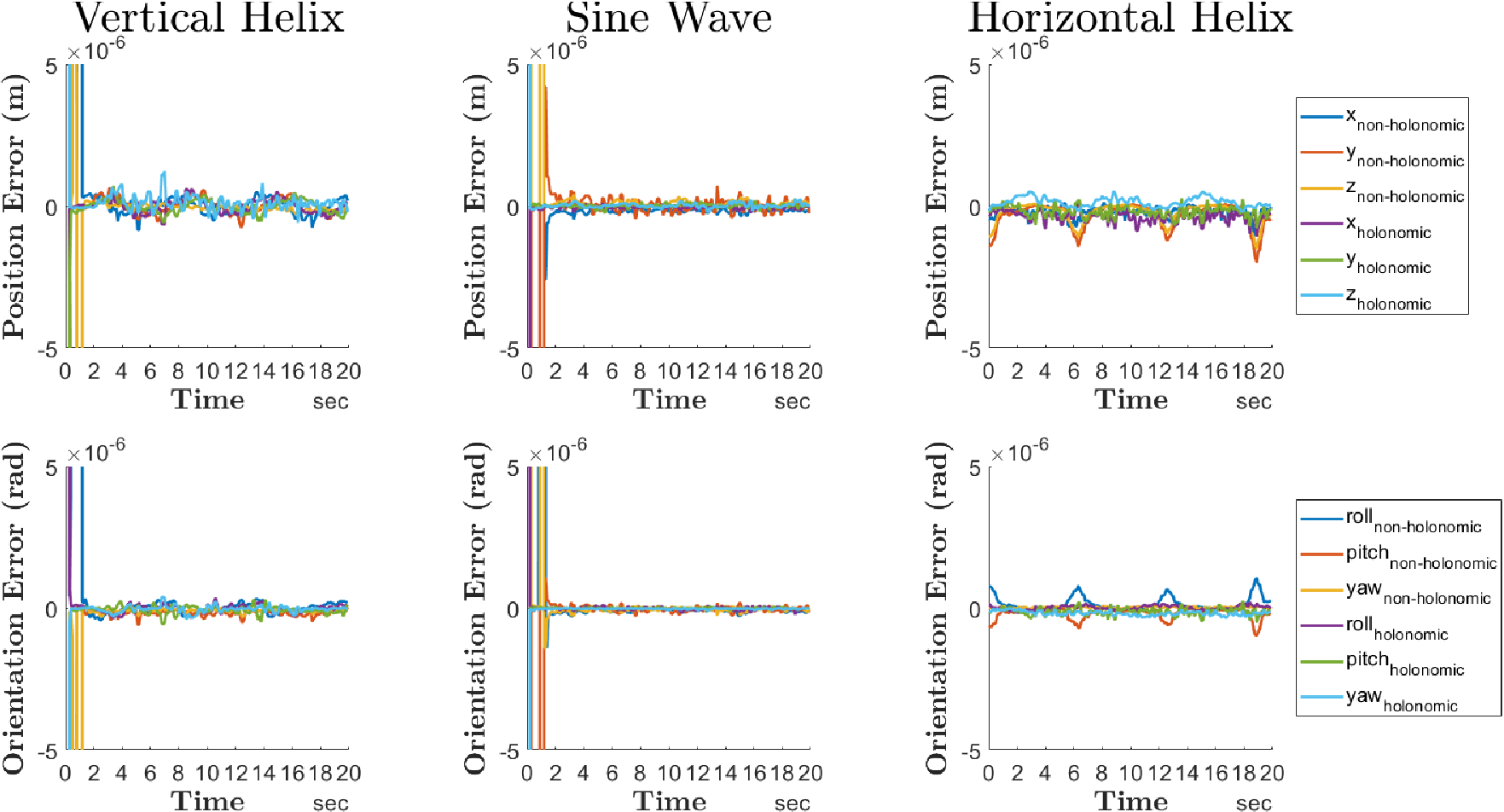}}
\caption{The error of the end effector on each of the different paths for both the position and orientation when using a non-holonomic and holonomic mobile platform.} \label{fig:error}
\end{figure}

The results of the experiment show that the proposed method is a successful inverse kinematic solver, suitable for a variety of robot and path configurations. 
The error for the different test paths is shown in Figure~\ref{fig:error}, with both the position and orientation error approaching zero. 
The average position and orientation \gls*{rmse} for the non-holonomic base were measured to be $0.0248$ meters and $0.0130$ radians, respectively. 
The mobile manipulator with a holonomic base performed similarly by accurately achieving the poses with a linear and angular error of $0.0237$ meters and $0.102$ radians, respectively. 

\begin{figure}[tbh]
\centerline {\includegraphics[width=\linewidth*3/4]{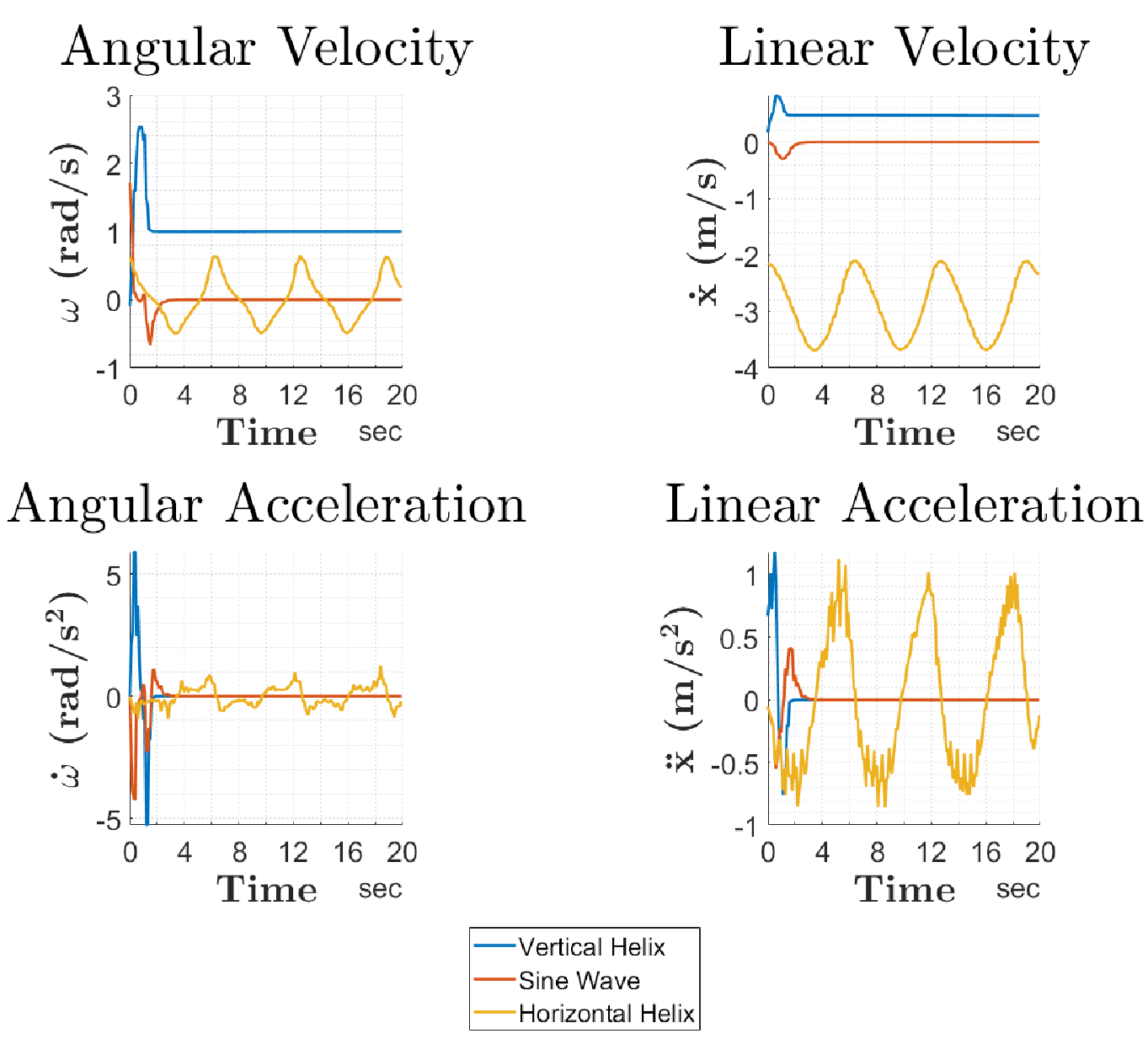}}
\caption{The computed angular and linear velocity for the mobile manipulator with a non-holonomic base.} \label{fig:full_base_nh}
\end{figure}

\begin{figure}[tbh]
\centerline {\includegraphics[width=\linewidth]{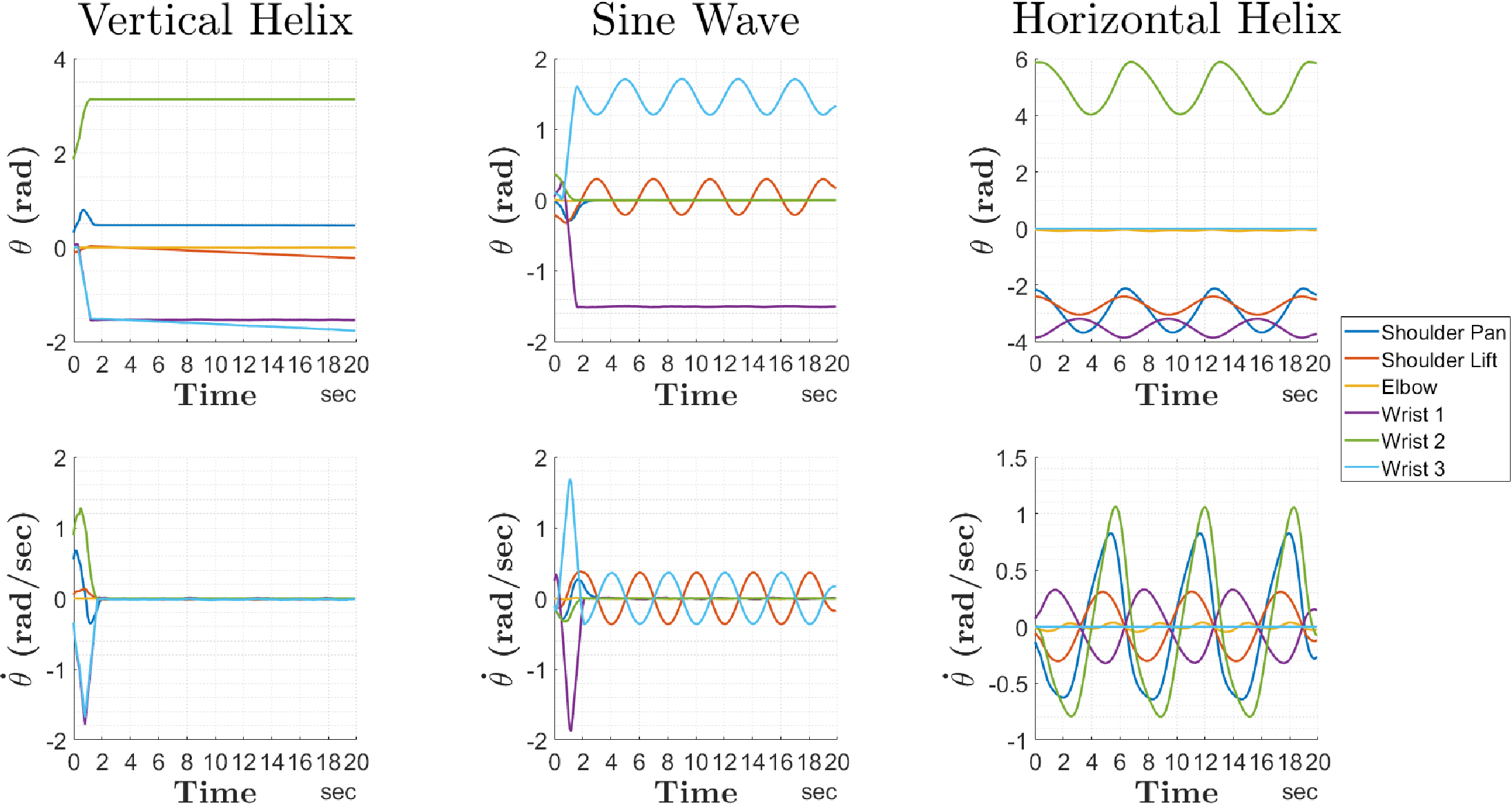}}
\caption{The computed joint states for the robotic manipulator mounted on the non-holonomic base.} \label{fig:full_joint_nh}
\end{figure}

\begin{figure}[tbh]
\centerline {\includegraphics[width=\linewidth]{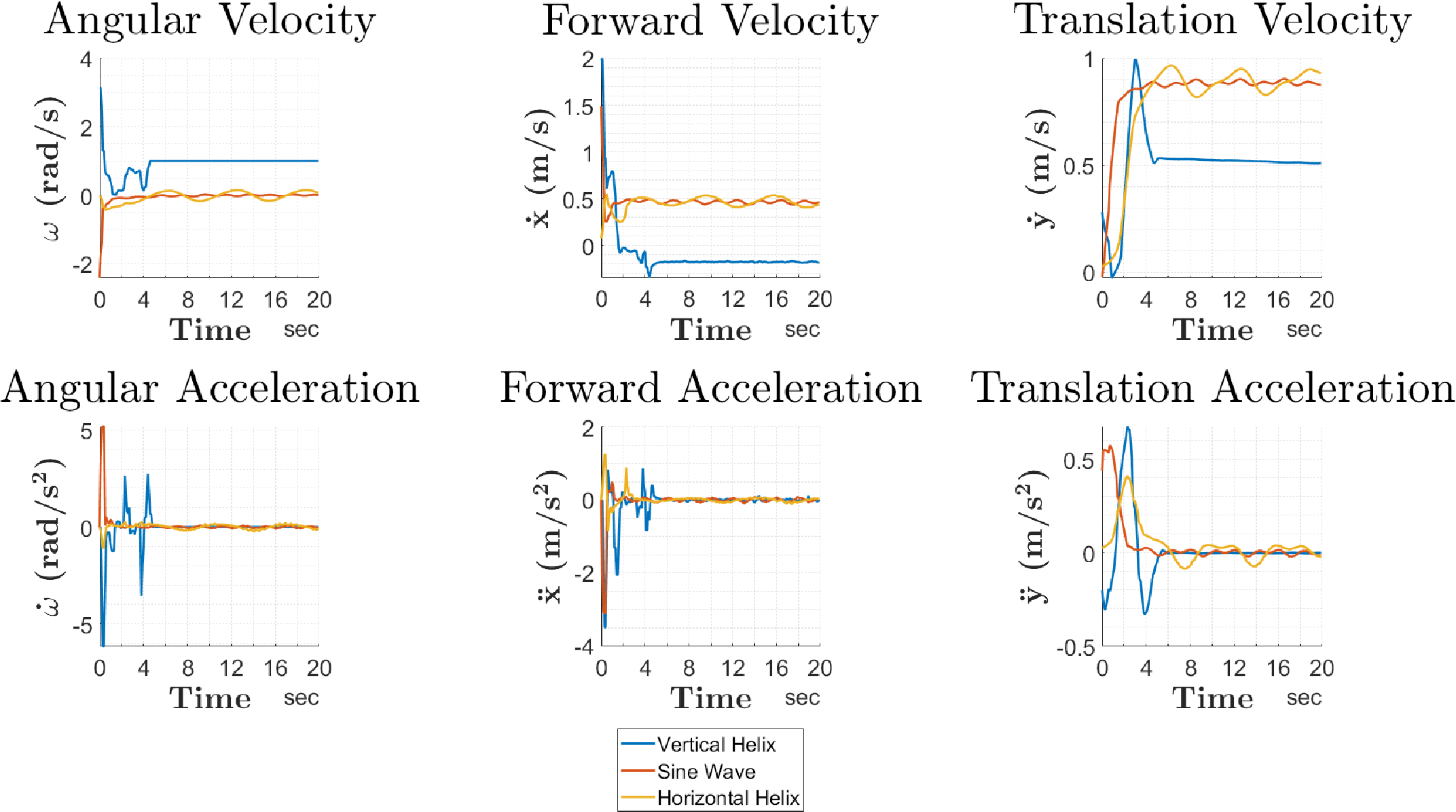}}
\caption{The computed angular and linear velocities for the mobile manipulator with a holonomic base. The forward and translational velocities represent the velocities in the x and y direction with respect to the base frame, respectively.} \label{fig:full_base_h}
\end{figure}

\begin{figure}[tbh]
\centerline {\includegraphics[width=\linewidth]{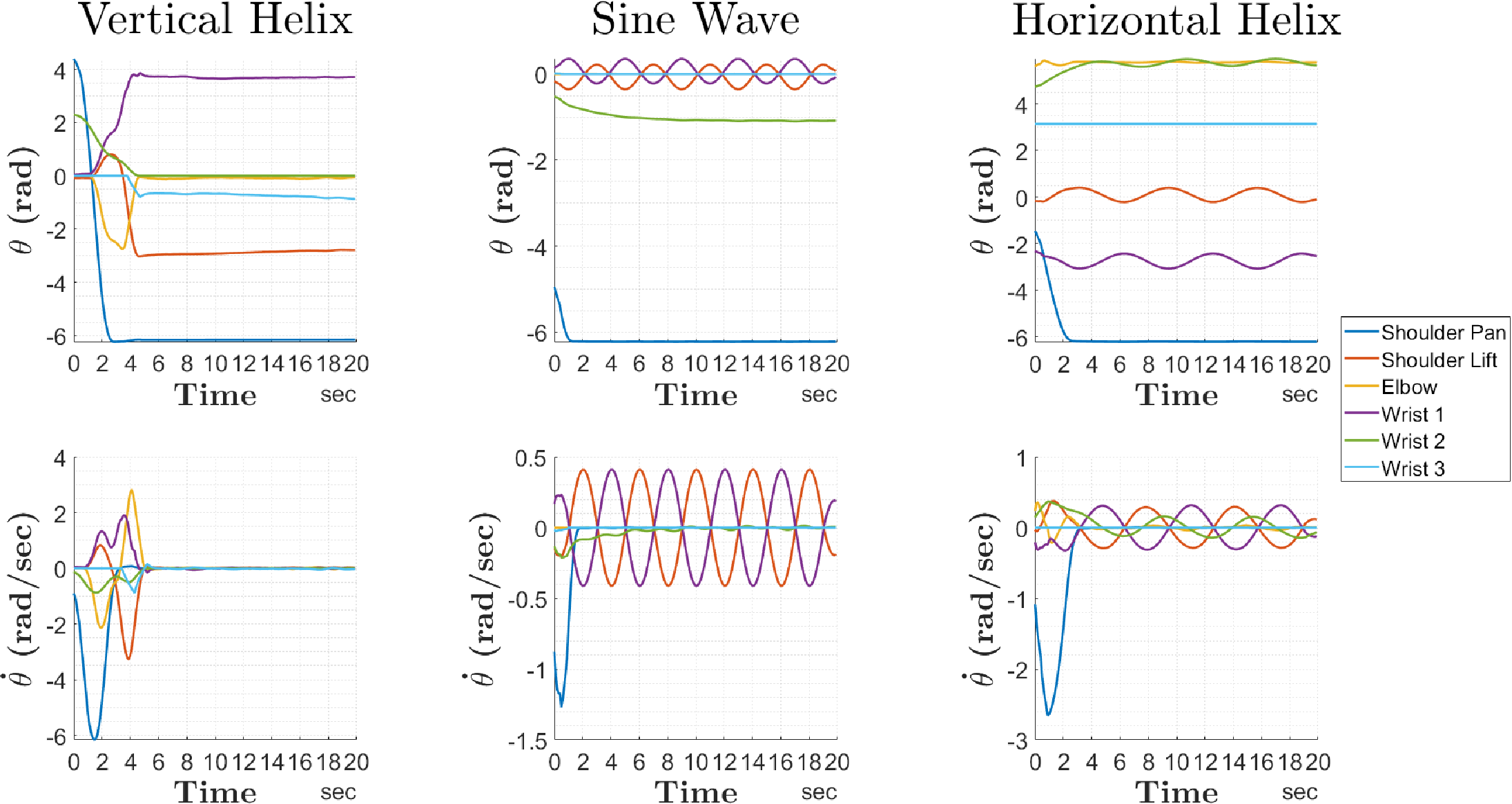}}
\caption{The computed joint states for the robotic manipulator mounted on the holonomic base.} \label{fig:full_joint_h}
\end{figure}

All the computed inverse kinematic variables are observed to be smooth and within reasonable bounds for a low level controller to execute.
The computed linear and angular velocities of the non-holonomic base are shown in Figure~\ref{fig:full_base_nh}. 
The optimized joint states of the manipulator on the non-holonomic base are shown in Figure~\ref{fig:full_joint_nh}. 
The pattern of the joint states compared to the motion of the base shows a unified motion plan for achieving the desired pose goals. 
The constraint weights contributed to the motion appropriately by limiting the change in motion for larger joints and allowing smaller joints to move sufficiently for finer corrections.
The holonomic base exhibits similar smoothness in the base and joint solutions, as seen in Figures~\ref{fig:full_base_h} and \ref{fig:full_joint_h}, respectively. 
The key metrics for performance validation on each of the paths are shown in Table~\ref{table:full_metric} for the non-holonomic and holonomic bases.
\begin{table*}[tbh]
\caption{Summarized metrics and results from the simulated experiment when testing the state planner on mobile manipulators.}
\label{table:full_metric}
\begin{center}
    \begin{tabular}{ | c | c c c | c c c | }
        \hline
        Mobile Base Configuration & & Non-holonomic & & & Holonomic & \\
        \hline
        Trajectories & Vertical Helix & Sine Wave & Horizontal Helix & Vertical Helix & Sine Wave & Horizontal Helix \\ 
        \hline
        Position RMSE (m) & 0.0415 & 0.0299 & 0.0031 & 0.0472 & 0.0147 & 0.0094\\  
        Rotation RMSE (rad) & 0.0214 & 0.0159 & 0.0016 & 0.0199 & 0.2841 & 0.0011\\  
        Computation Time (s) & 0.1862 & 0.1092 & 0.1216 & 0.1572 & 0.1075 & 0.1312 \\
        Max Forward Velocity (m/s) & 0.8317 & 0.3004 & 3.6881 & 2.0000 & 1.4923 & 0.5514\\ 
        Max Translation Velocity (m/s) & - & - & - & 0.9933 & 0.9028 & 0.9650\\
        Max Angular Velocity (rad/s) & 2.5299 & 1.7195 & 0.6449 & 3.1416 & 2.4128 & 0.4185\\
        Max Forward Acceleration (m/s$^2$) & 1.1774 & 0.5491 & 1.1238 & 3.4888 & 3.1030 & 1.2341\\
        Max Translation Acceleration (m/s$^2$) & - & - & - & 0.6746 & 0.5749 & 0.4107\\
        Max Ang Accel (rad/s$^2$) & 5.8892 & 4.2186 & 1.2413 & 6.1535 & 5.1969 & 1.0681\\
        Max Forward Jerk (m/s$^3$) & 5.8309 & 1.5568 & 3.4703 & 24.6442 & 15.3703 & 11.4400\\
        Max Translation Jerk (m/s$^3$) & - & - & - & 1.1581 & 0.6296 & 0.3900\\
        Max Angular Jerk (rad/s$^3$) & 23.7865 & 19.4592 & 6.7439 & 36.1791 & 25.6212 & 3.4932\\
        Max Joint Velocity (rad/s) & 1.7848 & 1.8732 & 1.0609 & 6.1663 & 1.2688 & 2.6576\\
        Max Joint Acceleration (rad/s$^2$) & 1.9905 & 2.5079 & 1.1399 & 5.4608 & 1.4042 & 1.8988\\
        Max Joint Jerk (rad/s$^3$) & 10.0259 & 7.9320 & 2.6727 & 9.6929 & 2.7737 & 3.4553\\
        \hline
    \end{tabular}
\end{center}
\end{table*}
The maximum values shown in the table, mostly come from the initial motion of the robot to achieve the first state and are otherwise significantly less.
The metrics and figures show the inverse kinematic planner can generate smooth and unified state curves for a mobile manipulator. 
Full results and graphics can be found at \url{https://github.com/peleito/slithers}.

Using Lie theory to formulate the objective function allowed for the high performance of the proposed method. 
Due to the kinematic constraints in the product of exponentials, the steady state error tends towards zero and the motion is smooth in between points. 
Since the objective function includes costs associated with sporadic motion, the state solutions are smooth between desired points. 
The combined kinematic model and optimization constraints have allowed the robotic system to show unified motion between the mobile base and manipulator.
The low error and smooth states show the benefits of using an optimization based approach to determine the desired states for the mobile manipulator. 
The performance of the proposed method on different mobile manipulators shows the method is generalizable for different robotic bases and manipulators. 
The results clearly show the capabilities of SLITHERS to plan the states for a mobile manipulator.  

\section{Conclusions}
\label{sec:conclusions}

This paper presented SLITHERS, a method for fast, accurate, and unified motion planning of mobile manipulators.  
The proposed method models the system using screw joints and Lie theory to define the forward and inverse kinematics. 
An objective function with smoothing constraints is then used to compute the optimized joint states to move the end effector from the current pose to the desired pose.

It was shown that the proposed inverse kinematics planner can generate feasible, smooth, and unified state curves. 
The error of the end effector pose compared to the desired pose is approaching zero upon reaching steady state. 
The base and joint velocities and accelerations are all within reasonable bounds and have no major discontinuities. 
Due to the evaluated performance, the proposed method is found to be a suitable state planner for mobile manipulators.

Future work includes testing the state planning performance on real world robots with noise and disturbances. 
The addition of position based force control using estimated deformations is being researched and pursued. 
Evaluation for planning with higher degrees of freedom systems, including humanoids, is sought to present in the upcoming opportunities including conferences and journals.






\bibliographystyle{IEEEtran}
\bibliography{references}

\end{document}